\documentclass{article}
\usepackage{spconf,amsmath,graphicx}

\usepackage{enumitem}
\setlist{nosep, leftmargin=14pt}

\usepackage{mwe} 
\usepackage[T1]{fontenc}
\usepackage{algpseudocode}
\usepackage{algorithm}
\usepackage{algpseudocode}
\usepackage{graphicx}
\usepackage{comment}
\usepackage{multirow}
\usepackage{booktabs}
\usepackage{subcaption}

\usepackage{amsmath}
\usepackage{amsfonts}
\usepackage{amssymb}

\newcommand{\bfsection}[1]{\vspace*{0.05cm}\noindent\textbf{#1.}}
\newcommand{\red}[1]{\textcolor{red}{#1}}

\newcommand{\etal}{\textit{et al}.}
\newcommand{\ie}{\textit{i}.\textit{e}.}
\newcommand{\eg}{\textit{e}.\textit{g}.}

\usepackage{xcolor}
\definecolor{orange}{rgb}{0.93, 0.53, 0.18}

\graphicspath{ {./images/} }
\usepackage[pagebackref=true,breaklinks=true,letterpaper=true,colorlinks=false,bookmarks=false]{hyperref}
\hypersetup{
     colorlinks = true,
     linkcolor = red,
     anchorcolor = black,
     citecolor = black,
     filecolor = black,
     urlcolor = black
     }
\title{S$^3$-TTA: Scale-Style Selection for Test-Time Augmentation in Biomedical Image Segmentation}

\name{Kangxian Xie$^{1,2}$\sthanks{The first author performed work during research internship at Boston College.} \quad Siyu Huang$^{3}$ \quad Sebastian Cajas Ordonez$^{4}$ \quad Hanspeter Pfister$^{4}$ \quad Donglai Wei$^{2}$}
\address{1. Boston University \quad 2. Boston College \quad 3. Clemson University \quad 4. Harvard University }

\begin{document}
%

\maketitle
\begin{abstract}
Deep-learning models have been successful in biomedical image segmentation. To generalize for real-world deployment, test-time augmentation (TTA) methods are often used to transform the test image into different versions that are hopefully closer to the training domain. Unfortunately, due to the vast diversity of instance scale and image styles, many augmented test images produce undesirable results, thus lowering the overall performance.
This work proposes a new TTA framework, S$^3$-TTA, which selects the suitable image scale and style for each test image based on a transformation consistency metric.
In addition, S$^3$-TTA constructs an end-to-end augmentation-segmentation joint-training pipeline to ensure a task-oriented augmentation.
On public benchmarks for cell and lung segmentation, S$^3$-TTA demonstrates improvements over the prior art by 3.4\% and 1.3\%, respectively, by simply augmenting the input data in testing phase.
\end{abstract}

\section{Introduction}
Segmentation is central to biomedical image analysis~\cite{hollandi2020nucleaizer}, which generates object masks, \eg, cell or lung segments, for downstream statistical and morphological analysis. 
However, learning-based methods often do not generalize well to unseen test images due to their mismatch in morphology and appearance with training ones.
To improve domain generalization and prevent test performance degradation, test-time augmentation (TTA) is a popular approach~\cite{style-invariant-cardiac,Moshkov2020TesttimeAF} to adapt the segmentation model to different domains.

\begin{figure}[t]
 \centering
\includegraphics[width=\linewidth]{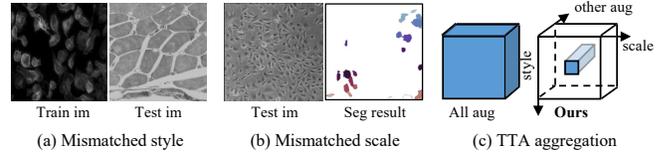}
\caption{\textbf{Scale-and-style-aware TTA.} (a-b) Due to the diversity of biomedical images, a pre-trained segmentation model may fail significantly if the test image has an unexpected style or scale. (c) Thus, instead of aggregating over all augmentations, we propose to select the suitable style and scale before the aggregation.}\label{fig:motivation}
\end{figure}

However, due to the vast diversity of biomedical images~\cite{cellpose}, a mismatched augmentation in style or scale can lead to poor segmentation results (Fig.~\ref{fig:motivation}\red{a-b}). Thus, previous approaches suffer from aggregating results from all versions of the augmented test image (Fig.~\ref{fig:motivation}\red{c}).
In this work, we propose a new TTA framework, scale-style selection for test-time augmentation (\textbf{S$^3$-TTA}), to automatically select the suitable image scale and style for each test image.
S$^3$-TTA constructs a task-oriented, augmentation-segmentation joint-training pipeline for image segmentation.
A scale-style selection unit adopts a self-consistency metric to select the best augmentated image as an intermediate feature suited for the following segmentation.
We evaluate S$^3$-TTA on public benchmarks of two biomedical domains, \ie, cell and lung segmentation, to demonstrate that it achieves the state-of-the-art performance.


\vspace{-2pt}
\subsection{Related Works}
\bfsection{Biomedical image segmentation}
Mask-RCNN~\cite{maskrcnn} takes the detect-to-segment approach and achieves decent performance for cells with simple shapes~\cite{Moshkov2020TesttimeAF,hollandi2020nucleaizer}.
For more complicated cell shapes, Cellpose~\cite{cellpose} creates a reversible mapping from vector flow and foreground prediction to an instance segmentation mask. 
For the out-of-the-domain setting, Lauenburg \etal~\cite{lauenburg2022instance} combines image translation and instance segmentation as a unitary design, while Keaton \etal
~\cite{keaton2023celltranspose} proposes a contrastive learning-based domain adaptation approach to adapt the pre-trained model. Distinct from these works, this paper focuses on adapting images in testing phase, where the domain adaptation methods are inapplicable.


\bfsection{Test-time augmentation (TTA) for segmentation}
TTA aims to improve the model performance on test image by augmenting it with geometric and appearance changes and then aggregating all results as a prediction. 
Moshkov\etal~\cite{Moshkov2020TesttimeAF} adopts geometric transformations, while Huang~\etal~\cite{style-invariant-cardiac} employs style transfer for appearance modifications.  Instead of aggregating predictions from all augmented samples, this work selects a single effective one for segmentation.

\begin{figure}[t]
\includegraphics[width=\linewidth]{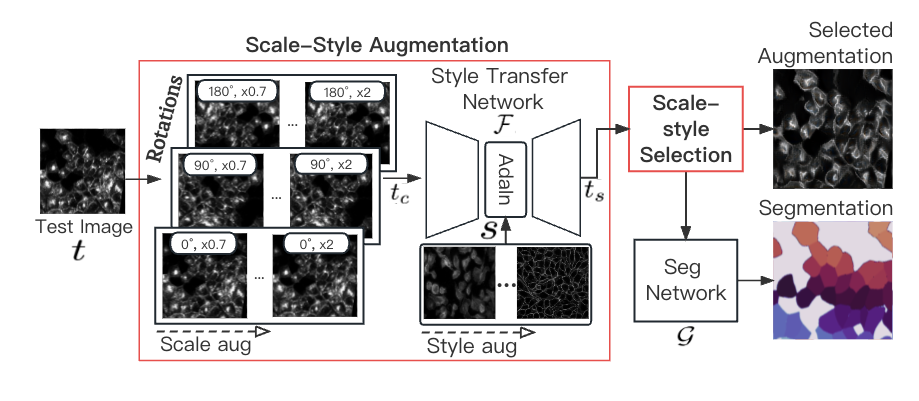}
\caption{\textbf{Model overview.} For a test image, we first apply scale, style augmentations at different angles. We then employ a consistency-based metric to select the best augmented images for segmentation.}
\label{fig:full}
\end{figure}
\bfsection{Image style transfer}
Gatys \etal~\cite{Gatys2015ANA} introduce a neural network model to transfer the artistic styles of images with a neural network. Subsequent works such as adaptive instance normalization (AdaIN)~\cite{adain} enable real-time and arbitrary style transfer. 
For data augmentation,~\cite{ADLFST,StyleAug,Li2020RandomST} adopts style transfer to generate new training images in other styles.
Specifically for test time augmentation, \cite{style-invariant-cardiac} transfers the styles of test images by randomly selecting styles from source data. Alternatively, Ma \etal~\cite{Ma2019NeuralST3D} use the Wasserstein metric for better style selection while Liu \etal~\cite{Liu2020RemoveAS} employ a feature histogram matching strategy. In addition, Liu \etal~\cite{Liu2021GeneralizeUI} formulates a 1-stage framework by directly conducting style transfer in segmentation with parameter injection.
This work also adopts neural style transfer for style augmentation and jointly trains the style transfer with the segmentation network to preserve segmentation-related details. 


%
%

\section{Method}
 

In this work, we propose an end-to-end test-time augmentation framework for image segmentation. As illustrated in Fig. \ref{fig:full}, the framework is composed of a scale-style augmentation module, a scale-style selector, and a U-Net-based segmentation network. The scale-style augmentation module resizes the input to different scales and applies style transfers as visual adaptation for task-oriented domain alignment. Then, the scale-style selector based on rotational-consistency picks the best augmented version for instance-segmentation. In the following, we discuss the proposed framework in detail.

\subsection{Scale-Style Image Augmentation}
\label{subsec:ss_augment}
We propose applying combinations of augmentations to the input, including scale augmentation and style augmentation, at multiple rotated versions (Sec. \ref{subsec:ss_select}) of the original image.

For scale augmentation, we adopt a series of different scaling-up ratios, e.g., $\times$1.5 and $\times$2, which rescale the input image into different spatial sizes. 
For style augmentation, we adopt a parametric method dubbed neural style transfer. More specifically, we adopt AdaIN~\cite{adain} as the style transfer network $\mathcal{F}$ due to its lightweight and adaptability to arbitrary styles. Take $t_{c}$ as image content and $s$ as style input, AdaIN operation is formulated as
\begin{equation}
\mathcal{F}(t_{c},s)= \sigma(s)*(\frac{t_{c}-\mu(t_{c})}{\sigma(t_{c})}) + \mu(s).
\end{equation}
In short, the ordered augmentation procedure involves rotations, resizing, and style transfer, forming a series of the input augmented to different versions, at different angles.  

\subsection{Scale-Style Selector}
\label{subsec:ss_select}

\begin{figure}[t]
\includegraphics[width=\linewidth]{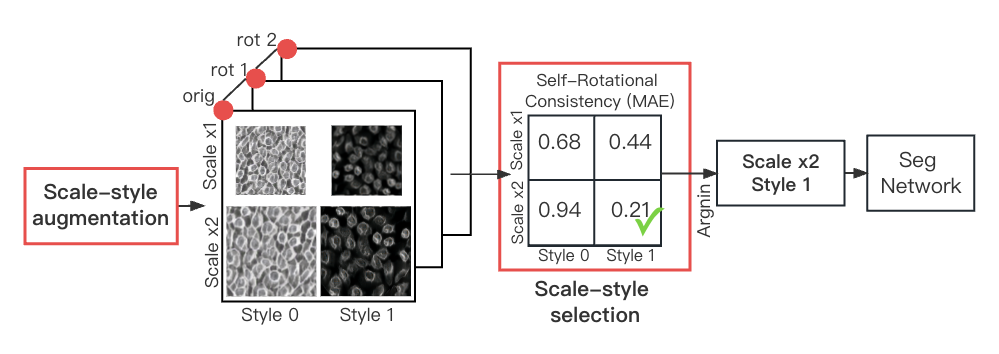}
\caption{\textbf{Scale-style selector.} This module measures the rotational-consistency of all augmentations and picks the best for segmentation. Two scales/styles are used for illustration.}
\label{fig:selection}
\end{figure}

While style transfer reduces the visual gap, it damages the content given unexpected style at the wrong scale. As Fig.~\ref{fig:motivation} shows, the segmentation network may fail to recognize images augmented to improper styles or scales. Therefore, instead of aggregating over all augmented segmentation results (Fig. \ref{fig:motivation}\red{c})\cite{Moshkov2020TesttimeAF,style-invariant-cardiac}, in this paper, we propose to select one augmentation policy from all combinations of augmentations.

More specifically, we devise a consistency-based scale-style selector. As illustrated in Fig. \ref{fig:selection}, for each augmentation policy, the transformation is conducted on the image at different angles. We measure its self-rotational-consistency on the image across multiple angles based on mean absolute error (MAE). Essentially, the augmented images at different angles are rotated back to original orientation, for pair-wise MAE measurements. The augmentation with the lowest averaged MAE, \ie, best consistency, is selected to proceed with the subsequent pipeline. The assumption behind this method is that the stylized image would be derandomized if the scale and style of cells is properly recognized by style transfer, and as the ST and segmentation are trained end-to-end (Sec. \ref{subsec:opt_inf}), they establish recognition on similar domain, which hints the plausibility of an augmentation policy for segmentation. This method is non-parametric and task-agnostic that it can be easily incorporated into existing TTA frameworks. 


\subsection{Optimization}
\label{subsec:opt_inf}
Prior to training, the ST network $\mathcal{F}$ is pre-trained on random content-style image pairs for training stability. We adopt the Cellpose model~\cite{cellpose} as the segmentation network $\mathcal{G}$. During training, the ST network encoder is fixed while the decoder is jointly trained with $\mathcal{G}$, which enables the following two benefits: (1) The style transfer network will be supervised by the boundary or foreground and background information to preserve more segmentation-related content details;
(2) The segmentation network will be finetuned on the style-transferred images to further narrow the visual gap between the training and test phases. Additionally, we incorporate the selection process into training, where only the selected augmented image passes through segmentation and optimization.

For the segmentation network, we employ the conventional practice and optimize it using the Cross-Entropy loss, written as $\mathcal{L}_{Seg}$.
The style transfer network is pre-train, and subsequently jointly-trained under content and style loss. The content loss computes MSE between the AdaIN output and the embedded feature of the stylized image. Let stylized image be $t_{s}$ and style input be $s$.
The content loss is formulated as
\begin{equation}
\mathcal{L}_{c} = ||\mathcal{F}(t_{s},s)-t||.
\end{equation}
The style loss measures the layer-wise difference in mean and variance between the style image and the stylized output. Let $\phi_{i}(x)$ be the output of the $i$-th ReLU layer of ST network $\mathcal{F}$, the style loss is written as
\begin{equation}
\mathcal{L}_{s} = \sum_{i=1}^{L} ||\mu(\phi_{i}(t))-\mu(\phi_{i}(s))|| + \sum_{i=1}^{L} ||\sigma(\phi_{i}(t))-\sigma(\phi_{i}(s))||
\end{equation}
The whole model is optimized with a weighted sum of the content loss, style loss, and segmentation loss with weights 1, 2, 5, respectively as $w_{c}$, $w_{s}$, and $w_{seg}$:
\begin{equation}
\mathcal{L}_{total} = \mathcal{L}_{Seg}*w_{seg}+(\mathcal{L}_{c}*w_{c}+\mathcal{L}_{s}*w_{s}),
\end{equation}




\begin{figure}[t]
\centering
\includegraphics[width=\linewidth]{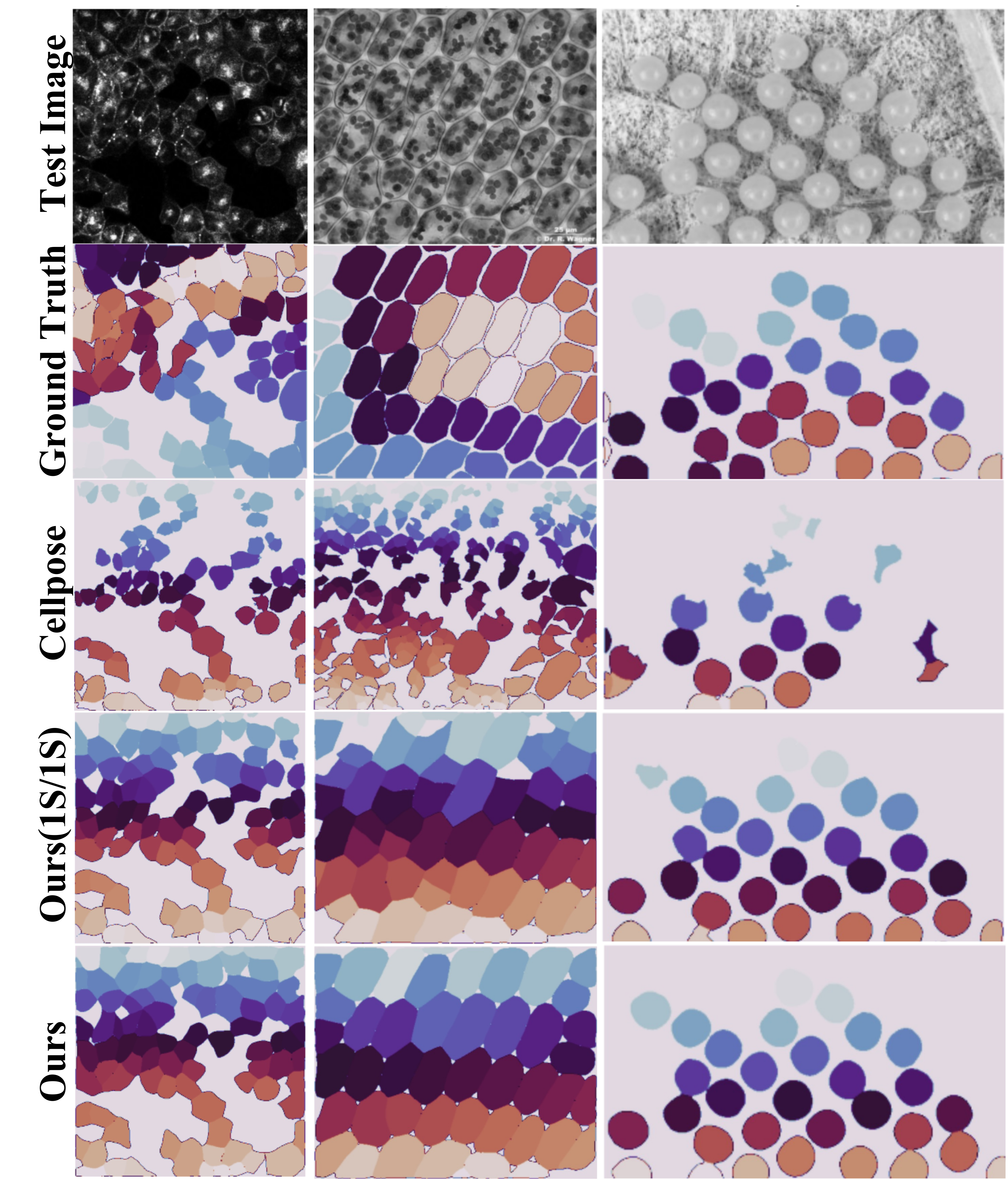}
\caption{\textbf{Qualitative results on cell image segmentation.} The fourth row represents our co-training pipeline without augmentation, while the final row presents results visualization with our multi-scale/style method.}
\label{fig:predictions_2}
\end{figure}
\vspace{-20pt}

\begin{table}[t]
\caption{Benchmark results on cell segmentation with Cellpose\cite{cellpose} as baseline, using F1 score (\%) with IOU matching thresholds $.5$, $.6$, and $.7$.}
\centering
\resizebox{1\linewidth}{!}{
\begin{tabular}{cc|ccc|cc}
\hline
 &  & Baseline\cite{cellpose} & TTA\cite{Moshkov2020TesttimeAF} & StyleInv\cite{style-invariant-cardiac} & Ours (1S/1S) & Ours \\ \hline
\multirow{3}{*}{CP-Full\cite{cellpose}} & .5 & 78.8 & 79.2 & 71.4 & 81.7 & \textbf{81.9} \\
 & .6 & 74.1 & 74.9 & 67.8 & 77.4 & \textbf{77.9} \\
 & .7 & 66.9 & 67.5 & 59.5 & 67.7 & \textbf{68.8} \\ \hline
\multirow{3}{*}{CP-Hard\cite{cellpose}} & .5 & 65.6 & 66.3 & 63.8 & 75.0 & \textbf{76.9} \\
 & .6 & 59.9 & 60.8 & 57.4 & 69.6 & \textbf{70.3} \\
 & .7 & 52.6 & 53.4 & 49.8 & 58.6 & \textbf{59.3} \\ \hline
\multirow{3}{*}{DSB2018\cite{dsb2018}} & .5 & 71.2 & 72.5 & 68.0 & \textbf{80.6} & 79.9 \\
 & .6 & 60.9 & 61.6 & 58.8 & \textbf{73.0} & 69.4 \\
 & .7 & 44.0 & 44.9 & 41.5 & \textbf{56.7} & 52.5 \\ \hline
\end{tabular}
}
\label{table:cell}
\end{table}

\section{Experiment}

\begin{figure}[t]
\centering
\includegraphics[width=\linewidth]{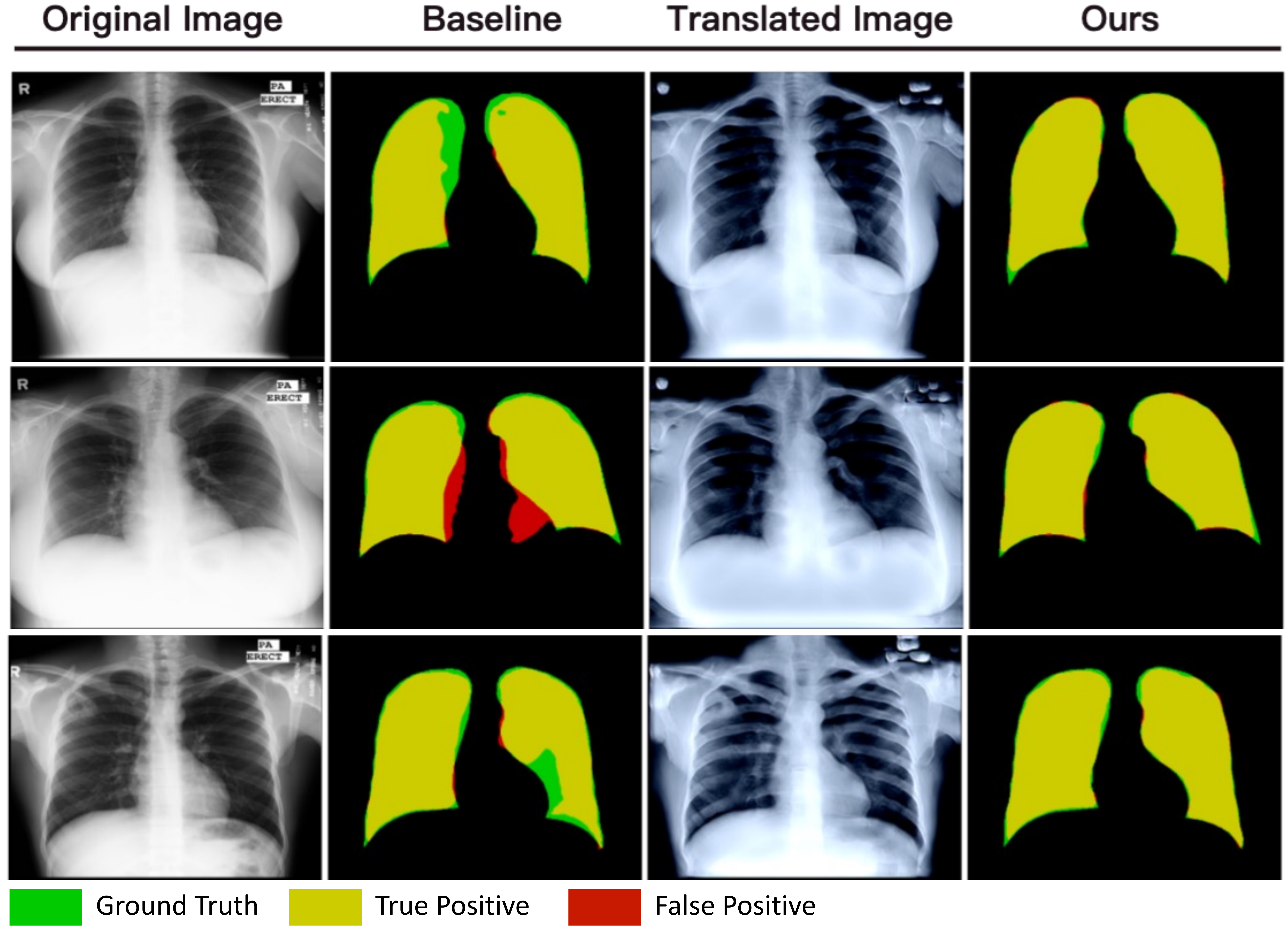}
\caption{Qualitative results on X-ray lung segmentation.}
\label{fig:predictions}
\end{figure}

\subsection{Datasets and implementation details}

\bfsection{Cell image benchmarks}
We evaluate our proposed framework with datasets in two tasks: Instance segmentation for microscopy cell images and semantic segmentation for chest x-ray lung segmentation. We apply instance-segmentation evaluation on the following 3 datasets:
(1) \underline{CP-Full}: The full testset (68 images) of the Cellpose dataset~\cite{cellpose}, a visually diverse multi-centered instance segmentation dataset with 616 cell images.
(2) \underline{CP-Hard}: We observed that 38 of the 68 Cellpose~\cite{cellpose} test images are in the training domain, i.e. from the same subset of training images. The remaining 30 images are from un-exposed domains. To assess the model's performance on unseen domains only, we evaluate the 30 images independently, which formulate the Cellpose-Hard Testset.
(3) \underline{DSB2018}~\cite{dsb2018}: Data Science Bowl 2018 presents a cell nuclei instance segmentation dataset with 670 training samples. We directly test methods on the training set, as it has mask labels for cell instances.  

\bfsection{X-ray image benchmarks}
We test our framework for chest X-ray lung segmentation. We adopted 3 chest X-ray datasets for evaluation:
(1) \underline{Shenzhen}~\cite{2cxr}: Collected by Shenzhen No.3 People’s Hospital, China, containing 662 frontal chest X-rays.
(2) \underline{Montgomery}~\cite{2cxr}: The dataset has 138 frontal chest X-rays.
(3) \underline{Darwin}~\cite{darwin}: The Darwin dataset is a sizable collection of 6106 diverse X-ray images.

\begin{table}[t]
\centering 
\caption{Benchmark results on X-ray lung segmentation, where the metric is Dice score/Jaccard index in percentage (\%).}
\resizebox{0.8\linewidth}{!}{
\begin{tabular}{l|cc|cc}
\hline
Train & \multicolumn{2}{c|}{Shenzhen} & \multicolumn{2}{c}{Montg.} \\ 
Test & Darwin & Montg. & Darwin & Shenzhen \\ \hline
Baseline\cite{deeplabv3} & 84/73 & 96/93 & 88/79 & 97/95 \\
Ours & \textbf{86/76} & \textbf{98/96} & \textbf{90/80} & 97/95\\ \hline
\end{tabular}
}
\label{table:xray}
\end{table}

\subsection{Benchmark results}

\bfsection{Cell image results}
We present quantitative comparison results on three cell instance segmentation datasets in Table \ref{table:cell}. We apply the F1 score as an evaluation metric, representing the ins-segmentation quality with multiple IOU matching thresholds. We compare our methods (1 scale/style and multi-scale/style) with the original Cellpose model and 2 baseline methods: test-time augmentation~\cite{Moshkov2020TesttimeAF} and style invariance~\cite{style-invariant-cardiac}. As a result, our pipeline outperforms all baselines in F1 score metrics with all IOU thresholds.
We show the qualitative results of cell instance segmentation in Fig. \ref{fig:predictions_2}. Our method contributes to better signal recognition as well as border matching.

\bfsection{X-ray image results}
We present results on chest x-ray lung semantic segmentation task. We employ Deeplab-V3~\cite{deeplabv3} as the Baseline network. 
As shown in Table \ref{table:xray}, our method performs best in both metrics. 
We show the qualitative results of lung semantic segmentation in Fig. \ref{fig:predictions}. The lung areas in the image after task-oriented style transfer are more prominent. Therefore, the task-oriented style transfer not only aligns visual differences between source and target domains but also helps accentuate the region of interest.  



\begin{table}[t]
\caption{F1 score (\%) with matching threshold $0.5$ on CP-Full, with different sets of scale and style augmentations.}
\centering
\resizebox{0.95\linewidth}{!}{
\begin{tabular}{c|cccc}
\hline
\# of Styles/\# of Scales & \{1\} & \{1,2\} & \{1,1.5,2\} & \{0.7,1,1.5,2\} \\ \hline
1  & 78.2 & 80.1 & 81.3 & 81.7 \\
3 & 79.3 & 81.0 & 81.6 & 81.9\\\hline
\end{tabular}
}
\label{table:ablation}
\end{table}

\subsection{Ablation studies and Analysis}

\bfsection{Number of Scale and Style}
Style input and scaling options compose our entire augmentation space. We separately evaluate the effectiveness of scale and style on Cp-Full with experiments on different numbers of scale-style options and present results in F1 score. Table \ref{table:ablation} shows that both components contribute to performance improvement. While incorporating more styles results in incremental F1 score gain, scaling options greatly improve results on images with small cells.

\bfsection{Style embedding visualization}
In Fig.~\red{6}, we visualize the distributions of Cellpose test images before and after style transfer in VGG-19 embedded space. Compared to the original images, the stylized images are located in a more condensed area. It demonstrates the effectiveness of style transfer, \ie, aligning the visual styles of images to smaller domains to enhance the segmentation performance.

\begin{figure}[t]
\centering
\includegraphics[width=0.9\linewidth]{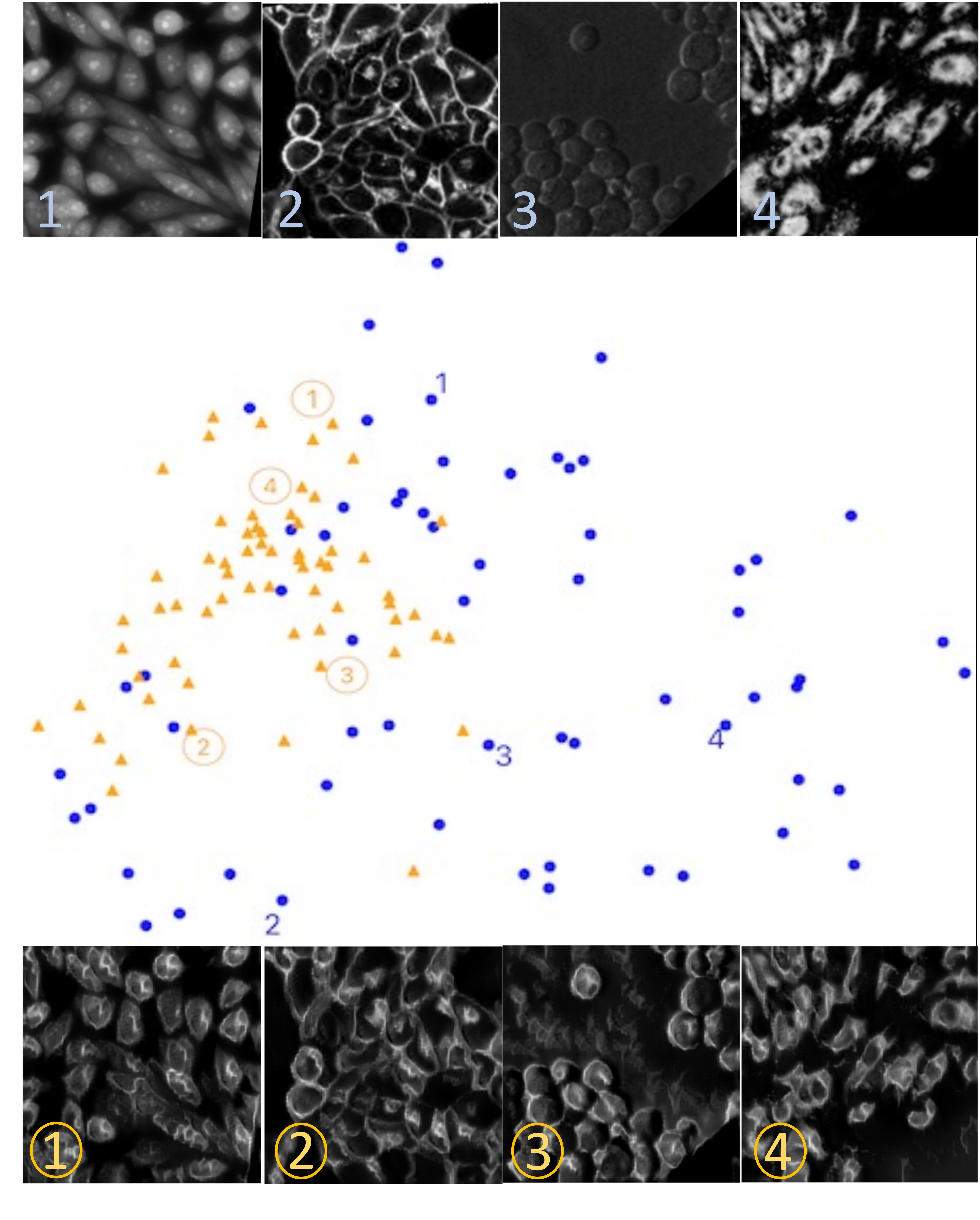}
\label{fig:viz}
\vspace{-12pt}
\caption{VGG feature embedding for original (top, \textcolor{blue}{blue}) and stylized images (bottom, \textcolor{orange}{orange}).}
\end{figure}
\section{Conclusion}
In this paper, we have proposed a novel TTA framework, dubbed S$^3$-TTA, to select the suitable style and scale for test images with a consistency metric. 
Furthermore, S$^3$-TTA consists of an end-to-end augmentation-segmentation training pipeline to ensure a task-oriented augmentation. In two biomedical image domains, the proposed framework significantly outperforms the prior art.

\section{Compliance with ethical standards}
This research study was conducted retrospectively using open access human subject data from MC set~\cite{2cxr} and Shenzhen set~\cite{2cxr}, made available by the U.S. National Library of Medicine. Ethical approval was not required.

\section{Acknowledgment}
This research is supported in part by the NSF award IIS-2239688.
\bibliographystyle{IEEEbib}
\bibliography{refs}
\end{document}